\documentclass[10pt,twocolumn,letterpaper]{article}

\usepackage[pagenumbers]{cvpr}      

\usepackage[dvipsnames]{xcolor}
\usepackage{multirow}
\usepackage{booktabs}
\usepackage{algorithm}
\usepackage{algorithmic}
\usepackage{amsmath}
\usepackage{pifont}
\usepackage{xcolor}
\usepackage{graphicx}
\usepackage{amsmath}
\usepackage{pifont}
\usepackage[table]{xcolor}
\usepackage{subcaption}
\usepackage{tabularx}

\definecolor{cvprblue}{rgb}{0.21,0.49,0.74}
\usepackage[pagebackref,breaklinks,colorlinks,allcolors=cvprblue]{hyperref}

\def\paperID{***} 
\def\confName{CVPR}
\def\confYear{2025}

    \title{ViewBridge: Curriculum Knowledge Distillation for Activity View-Invariance Under Extreme Viewpoint Changes}

\author{
Arjun Somayazulu\textsuperscript{1}\hspace{0.2cm}
Efi Mavroudi\textsuperscript{2}\hspace{0.2cm}
Changan Chen\textsuperscript{3}\hspace{0.2cm}
Lorenzo Torresani\textsuperscript{4}\hspace{0.2cm}
Kristen Grauman\textsuperscript{1}\hspace{0.2cm}\\[0.3cm]
\textsuperscript{1}UT Austin \hspace{0.2cm}
\textsuperscript{2}Meta AI \hspace{0.2cm}
\textsuperscript{3}Stanford University \hspace{0.2cm}
\textsuperscript{4}Northeastern University
}
\begin{document}
\maketitle
\begin{abstract}
Traditional methods for view-invariant learning rely on controlled multi-view training data with minimal scene clutter. However, they struggle with in-the-wild videos that exhibit extreme viewpoint differences and share little visual content. We introduce \textbf{ViewBridge}, a framework for learning rich video representations in the presence of severe view-occlusions. We introduce a knowledge distillation objective that preserves action-centric semantics, together with a novel curriculum learning procedure that pairs incrementally more challenging views over time, thereby allowing smooth adaptation to extreme viewpoint differences. To sort training video segments for the proposed curriculum, we define a geometry-based metric that reflects their likely occlusion level.  While training leverages multi-view data, at inference time, the input is an uncalibrated, single-viewpoint video. Evaluating our approach on two tasks---temporal keystep grounding and fine-grained keystep recognition---we outperform SOTA approaches across three datasets (Ego-Exo4D, LEMMA, EPFL-Smart-Kitchen-30).\footnote{Project page: \url{https://vision.cs.utexas.edu/projects/learning_view_distill/}}
\end{abstract}    
\section{Introduction}
\label{sec:intro}

\begin{figure*}[t]
  \centering   \includegraphics[width=1.0\linewidth]{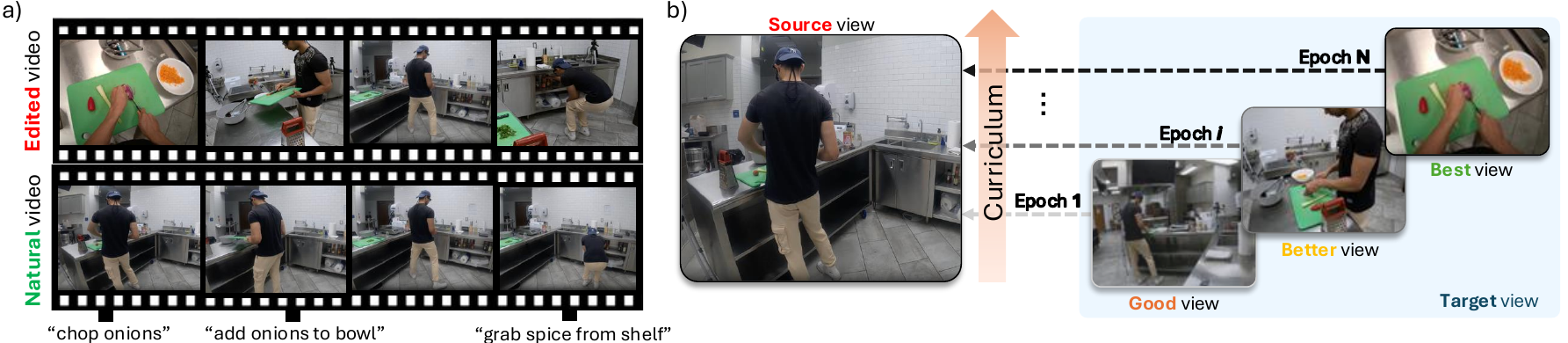}
   \caption{\textbf{Edited vs. natural procedural video.} \textbf{a)} Whereas edited video switches between close-in  and wide-body shots to best capture the ongoing action, natural in-the-wild video instead experiences significant object and view occlusions. \textbf{b)} Directly distilling the best view into an impoverished viewpoint has limited utility, given the lack of shared visual content. ViewBridge aligns features from source views with an \emph{incrementally better target view} that still shares significant visual content. As training proceeds, we incorporate target views that better capture the ongoing action, but share less visual similarity with the source view.}
    \label{fig:concept_fig}
\end{figure*}

Across a wide range of everyday human activities, certain viewpoints better capture ongoing actions than others. In an activity involving object interactions such as cooking, the first-person view showcases the ingredients, utensils, and fine-grained hand movements of a chef as they follow a recipe; in arts and crafts, household chores, DIY tasks, repair tasks, first aid, and many others, the situation is similar. Meanwhile, exocentric (exo) views of the same activities can reveal the subject's body pose and wider scene context. However, the visibility that each view provides is highly context-dependent and changes over time during the execution of an activity (Fig.~\ref{fig:concept_fig}).

Understanding actions from arbitrary \emph{single-view} video is essential to handling in-the-wild data, yet the edited nature of many popular datasets has allowed the research community to skirt this challenge. Existing work in action recognition and temporal grounding relies on video(-text) datasets curated from YouTube or movies \cite{miech2019howto100mlearningtextvideoembedding, NEURIPS2023_9d58d85b, caba2015activitynet,soldan2022madscalabledatasetlanguage,kay2017kineticshumanactionvideo, carreira2018shortnotekinetics600,soomro2012ucf101dataset101human, 6126543}, which consist of stylized, studio-edited videos that intelligently shift between shots to ensure an optimal view of the activity at each moment. Similarly, popular single-view and multi-view video data is typically collected in controlled settings with minimal clutter, intelligent camera placement, or explicit actor instructions to ensure that actions are always visible and centered \cite{tacos:regnerietal:tacl,10.1016/j.cviu.2006.07.013, shahroudy2016ntu, liu2020ntu}. In contrast, in-the-wild activity  recorded passively from a static exo viewpoint lacks such editing and controls, so the subject's actions may frequently be partially occluded from view (Fig.~\ref{fig:concept_fig}, ``natural video").

One general strategy to cope with viewpoint variability is to learn \emph{view-invariant} representations that, ideally, would capture a stable representation of the video content, regardless of the viewpoint \cite{zhang2017viewadaptiverecurrentneural, 7298860, 9945457,sarkar2024learningstateinvariantrepresentationsobjects}.
Existing approaches seek to capture information \textit{shared among all views} via object \cite{NEURIPS2023_a845fdc3} or body-pose correspondences \cite{zhang2017viewadaptiverecurrentneural, 7298860, 9578793, 9945457, vipoem2020}. While this works well for certain activities having minimal objects, clutter, and occlusions (e.g., a basketball player on a empty court), the assumption of mutually shared content breaks down in more general cases. The issue is exacerbated in activities where hand-object interactions are significant to understanding. For example, a camera on the counter pointed at a person's active workspace will share little information with a camera mounted across the room facing the back of the subject.  Furthermore, as a subject moves through a space over the course of the activity, the visible content becomes more or less informative. In these common settings, the traditional approach of directly enforcing agreement between views is destabilizing; the lack of shared visual content yields weak representations. 

To address this challenge, we propose \textbf{ViewBridge}, a new approach to learning view-invariant visual representations from multi-view training data.  Our key insight is to overcome extreme visibility differences by \emph{gradually distilling information from more informative views into visually impoverished views}. We focus on activities featuring hand-object interactions---a wide class representing a frontier for video understanding.

Given multi-view synchronized training videos, we first introduce a geometry-based metric for ranking each viewpoint by its visibility of the region acted upon at each moment. We use these rankings to design a cross-view knowledge distillation objective that maximizes feature similarity between the input view and a view that better observes the ongoing action. To address the large viewpoint disparity between the best view and other occluded views, we further devise a curriculum learning strategy that aligns synchronous features from occluded views with incrementally higher quality views (Fig.~\ref{fig:concept_fig}, bottom). At inference, the input is standard \emph{single-view} exocentric video, whose features are enriched by our multi-view training process.

We tackle two complementary tasks for evaluation: fine-grained keystep recognition from trimmed video clips, and language-guided keystep grounding in untrimmed video---both tasks that suffer in the presence of poor visibility.  We demonstrate SOTA results across three challenging multi-view datasets (Ego-Exo4D~\cite{grauman2024egoexo4dunderstandingskilledhuman}, LEMMA~\cite{jia2020lemmamultiviewdatasetlearning}, EPFL-Smart-Kitchen-30~\cite{bonnetto2025epflsmartkitchen30denselyannotatedcooking}). Our results strongly support curriculum distillation as a powerful learning paradigm for unedited, in-the-wild activity video.
\section{Related Work}
\label{sec:related_work}

\paragraph{View-invariant video representations.} View-invariant representation learning has been explored more extensively for static images \cite{sarkar2024learningstateinvariantrepresentationsobjects, li2023learningdistortioninvariantrepresentation, doersch2016unsupervisedvisualrepresentationlearning, sardari2024unsupervisedviewinvarianthumanposture, kanezaki2018rotationnetjointobjectcategorization}, with only limited attention on video.  Existing methods learn view-invariant features for cross-view action recognition \cite{2011savaresecrossview, 10.1007/978-3-540-88682-2_13, Zhang_2013_CVPR, 7298860, 6123211, NEURIPS2023_a845fdc3,grauman2024egoexo4dunderstandingskilledhuman}, typically relying on time-synchronized multi-view video for training as now widely available~\cite{10.1016/j.cviu.2006.07.013,wang2014crossviewactionmodelinglearning,grauman2024egoexo4dunderstandingskilledhuman,jia2020lemmamultiviewdatasetlearning,sener2022assembly101largescalemultiviewvideo,bonnetto2025epflsmartkitchen30denselyannotatedcooking,khirodkar2024harmony4dvideodatasetinthewild}. 
These methods rely on features shared across views, whereas our approach explicitly targets scenarios where extreme viewpoint differences and clutter result in wider variation in cross-view shareable content.  No prior work explores explicit staging of training according to viewpoint, as we propose.

\paragraph{Ego-exo translation and transfer.}
Recent work explores ways to transfer information specifically between ego and exo
viewpoints, whether to enhance representations for action recognition \cite{Li2021EgoExoTV,ARDESHIR201861,sigurdsson2018charadesegolargescaledatasetpaired,NEURIPS2023_a845fdc3,grauman2024egoexo4dunderstandingskilledhuman,2014shapiroactionrecognition}, cross-view retrieval \cite{Yu2019WhatIS,ARDESHIR201861}, or new-view image synthesis \cite{luo2024shoesliftingegocentricperspective,10.1007/978-3-031-72691-0_23}. Note that methods that assume multi-view input at test time (e.g.,~\cite{2014shapiroactionrecognition}) are out of scope; we focus on single-view input due to its broader applicability. Related to these methods, we are also interested in how ego and exo views can be mutually influential.  However, whereas prior work treats all exocentric views uniformly \cite{10031025, siddiqui2023dvanetdisentanglingviewaction, 2011savaresecrossview}, our idea to guide training based on shared visibility is entirely new. Our curriculum learning objective explicitly assesses the quality of each view and uses it to smoothly adapt distillation between increasingly extreme viewpoints, targeting informative features from highly occluded video.

\paragraph{3D for robust activity recognition.}
Besides cross-view alignment, other work uses 3D body pose and human meshes for view
invariance \cite{zhang2017viewadaptiverecurrentneural, 7298860, 9578793, 9945457, 10.1016/j.cviu.2006.07.013, vipoem2020,george-cvpr2023}, particularly for action recognition from unseen viewpoints. In addition, extrinsic camera parameters \cite{10030791} and 3D flow \cite{NEURIPS2018_2f37d101} help learn self-supervised world-view-invariant video representations for action recognition. 
In contrast, ViewBridge uses geometry only as weak training-time supervision to rank views, enabling robust single-view activity understanding without 3D pose, calibration, or multi-view input at inference.
\section{ViewBridge}

We propose a training paradigm for learning view-invariant activity representations from multi-view synchronized un-edited videos. We first introduce a training objective that distills information from higher quality views into lower quality views 
(Sec.~\ref{sec:k_dist}), followed by a curriculum learning strategy that selects distillation targets from increasingly disparate viewpoints over the course of training (Sec.~\ref{sec:curr_training}). We then describe how we assess view quality and obtain a ranking of all views at each time step based on their geometric and semantic properties (Sec.~\ref{sec:camera_ranking}). Finally we introduce the downstream tasks and models (Sec.~\ref{sec:tasks}) in which we deploy ViewBridge.

\subsection{Cross-view knowledge distillation}
\label{sec:k_dist}
Whereas existing work learns information \emph{shared} across viewpoints, we seek to enrich features from impoverished or occluded viewpoints with information from viewpoints having \emph{better visibility} of ongoing actions and objects. Given a ranking of view quality (to be described in Sec.~\ref{sec:camera_ranking}), we design an objective that distills knowledge from higher quality views into poorer quality views.

\begin{figure*}[t]
  \centering
   \includegraphics[width=1.0\linewidth]{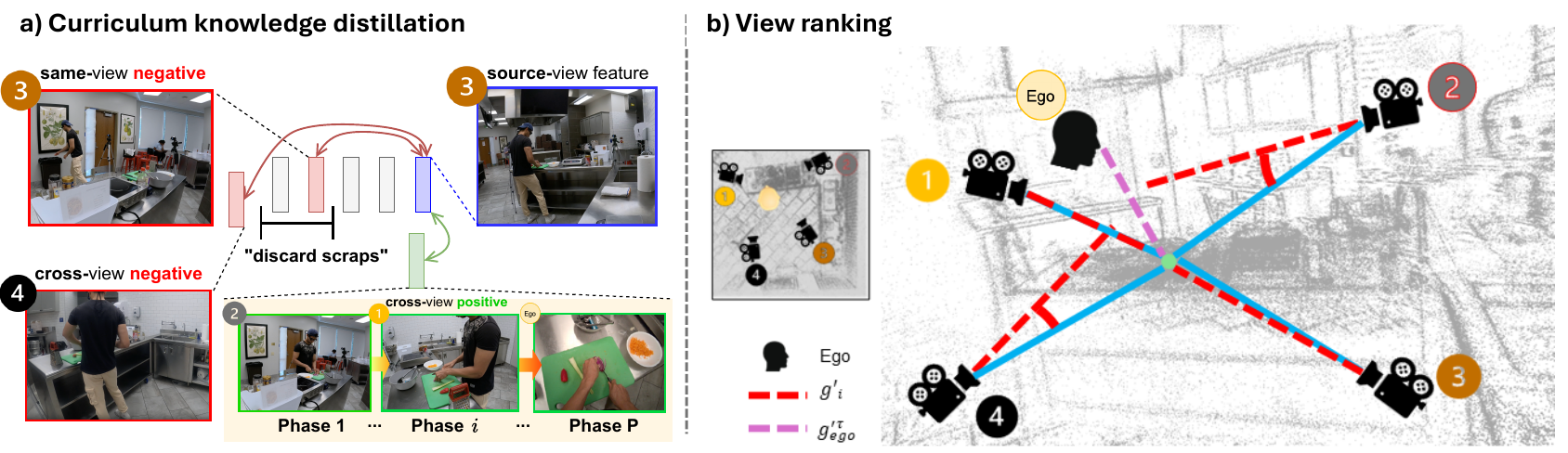}
   \caption{\textbf{ViewBridge overview.} \textbf{a)} For each feature from a source view (blue), we minimize similarity with the synchronous worst-rank view (cross-view negative) and with a feature from the same view demonstrating a different keystep (same-view negative). Our curriculum chooses a positive distillation target (cross-view positive) from incrementally higher-rank views over the course of training. See Sec.~\ref{sec:k_dist}/~\ref{sec:curr_training}. \textbf{b)} Given an ego-worn camera looking down at the active workspace, we rank each exo camera by its view-alignment with the hand-object interaction region $p_{\text{center}}$ (green). To account for self-occlusion by the camera-wearer, we enforce that views in front of the ego-camera (1, 2 in this example) are ranked ahead of views behind the ego-camera (3, 4). See Sec.~\ref{sec:camera_ranking}.
   }
    \label{fig:approach_fig}
\end{figure*}

\textbf{Cross-view distillation target.} Let $\mathcal{S} = \{v_{ego}\} \cup V_{exo}$ denote the synchronized set of views for a $T$-second multi-view video take, consisting of a single egocentric (``ego'') camera $v_{ego}$ and $N$ exocentric (``exo'') cameras $V_{exo}$ (see Fig.~\ref{fig:approach_fig} (b)). Such multi-view data is of increasing interest and availability~\cite{10.1016/j.cviu.2006.07.013,wang2014crossviewactionmodelinglearning,grauman2024egoexo4dunderstandingskilledhuman,jia2020lemmamultiviewdatasetlearning,sener2022assembly101largescalemultiviewvideo,bonnetto2025epflsmartkitchen30denselyannotatedcooking,khirodkar2024harmony4dvideodatasetinthewild} as researchers elevate activity understanding from a purely frame-level task to a spatially grounded problem. For each timestep $\tau$, let $R_\tau$ denote an ordering of the views in $\mathcal{S}$ by visibility of the ongoing action, and let $\rho_\tau(v)$ be the rank of view $v$ in this ordering, with lower ranks indicating better visibility of the ongoing activity. We  define $R_\tau$ in Sec.~\ref{sec:camera_ranking}. 

Given per-second video features $F^{v_i} = [f^{v_i}_1, \ldots, f^{v_i}_T]$ from source view $v_i$, for each timestep $\tau$, we choose the synchronous feature $f^{v_{pos}}_\tau$ from a higher-quality view $v_{pos}$ as a positive cross-view distillation target, where $v_{pos}$ is a higher quality view selected from the view ranking according to our curriculum strategy (Sec.~\ref{sec:curr_training}).
 
We also select $f^{v_w}_\tau$ as a negative target, where
\begin{equation}
    v_w = \arg\max_{v \in \mathcal{S}} \rho_\tau(v)
\end{equation}
is the view with poorest quality at time $\tau$. We dynamically vary the rank used to obtain $v_{pos}$ according to our curriculum strategy (Sec.~\ref{sec:curr_training}). See Fig.~\ref{fig:approach_fig} (a).

\textbf{Same-view negative sampling.} To ensure we capture action semantics and not view-specific information, we choose an additional negative feature target from the same view. Our distillation objective uses activity step descriptions (keysteps) and their temporal intervals (in the form of start/end timestamps) during training. For feature $f^{v_i}_\tau$ at time $\tau$, we first identify the text keystep that has minimal similarity with $f^{v_i}_\tau$ in a shared vision-text embedding space
\begin{align}
    j_{\text{neg}} = \arg\min_j \text{cos}(f^{v_i}_\tau, k_j),
\end{align}
where $\{k_j\}^K_{j=1}$ is the set of $K$ keystep features associated with the video. We then randomly sample a video feature $f^{v_i}_{t_{neg}}$ from within $k_{j_{\text{neg}}}$'s ground truth temporal interval $(s_{j_{\text{neg}}}, e_{j_{\text{neg}}})$, where $s_{j_{\text{neg}}}$ and $e_{j_{\text{neg}}}$ are the start/end timestamps for $k_{j_{\text{neg}}}$.

\textbf{InfoNCE loss.} We use the positive feature $Q = \{f^{v_{pos}}_\tau\}$ and negative features $G = \{f^{v_w}_\tau, f^{v_i}_{t_{neg}}\}$ in an InfoNCE loss: 
\begin{equation}
\mathcal{L}_{\text{InfoNCE}} = -\log \frac{\scriptstyle \sum\limits_{q \in Q} \exp(\text{sim}(q, f^{v_i}_\tau) / \gamma)}{\scriptstyle \sum\limits_{q \in Q} \exp(\text{sim}(q, f^{v_i}_\tau) / \gamma) + \sum\limits_{g \in G} \exp(\text{sim}(g, f^{v_i}_\tau) / \gamma)},
\end{equation}
where $\text{sim}(\cdot,\cdot)$ is cosine similarity, and $\gamma=0.1$ is the temperature. This loss aligns synchronous features from poor quality views with higher-quality views that better observe the activity across all timesteps.

\subsection{Viewpoint-driven curriculum}
\label{sec:curr_training}
Directly distilling information between views that share little visual content is a significant learning challenge. To address this, we propose a curriculum learning strategy that allows our knowledge-distillation objective to smoothly adapt to distillation between features with large viewpoint differences during training. When training the task models incorporating our ViewBridge framework (Sec.~\ref{sec:tasks}), 
we divide training into $P$ phases, where $P$ is the maximum number of views for any take (ego+exo). In each phase $p$, we choose the cross-view positive distillation target from view $v_{pos}$ using the source-excluded ranking $R_\tau^{\setminus v_i}$, obtained by removing the source view $v_i$ from $R_\tau$:
\begin{equation}
    v_{pos} =
    R_\tau^{\setminus v_i}\!\left[\max(0, \rho_\tau(v_i)-p)\right].
\end{equation}
Here $R_\tau^{\setminus v_i}[j]$ denotes the view at index $j$ in the source-excluded ranking. This ensures that the positive target is always a distinct synchronous view. In phase $p=1$, our curriculum aligns input-view features with their \textit{immediate next-best} ranked view as target, which may not optimally observe the activity, but shares large visual overlap with the input view. In phases $p > 1$, distillation targets are selected from views with incrementally better visibility of the ongoing action (Fig.~\ref{fig:approach_fig} (a)).

\subsection{Activity-centric view quality ranking}
\label{sec:camera_ranking}

The distillation objective above requires an ordering of views by how well they observe the ongoing activity. The ideal view quality ranking would move from views that have greatest visibility of the activity, to those that have the least. We hypothesize that the ego view from a head-mounted camera yields optimal visibility of any activity involving object interactions, since it observes the objects/hands at all times and maintains consistent visibility even as the subject moves about the scene. Therefore, we enforce that $v_{ego}$ is first in our ranking of each training take's camera views. At time $\tau$, we obtain a ranking of all views
\begin{equation}
    R_\tau = [v_{ego}, v^\tau_1,\ldots, v^\tau_N],
\end{equation}
where $v^\tau_1$ is the exo view with best visibility of the hand-object interaction region and $v^\tau_N$ is the exo view with poorest visibility at time $\tau$. The rank function $\rho_\tau(v)$ used above is induced by this ordering.

Our ranking of all the exo views is motivated by two factors affecting how informative and how easily ``linkable'' views are: 1) the extent of the shared hand-object interaction (HOI) region (semantic) and 2) the amount of occlusion (geometric).
Hence we rank according to a view's mutual visibility with the hand-object interaction region, as follows.

Given extrinsic camera parameters $K_i = [R_i, t_i]$ for exo camera $i$ and $K^\tau_{ego} = [R^\tau_{ego}, t^\tau_{ego}]$ for dynamic egocentric camera $v_{ego}$ at time $\tau$, we convert to the world-coordinate reference frame:
\begin{equation}
[R'|t'] = [R^T | -R^T\cdot t].
\end{equation}
We then estimate the center of the hand-object interaction region at time $\tau$ by projection along the ego camera orientation (viewing) vector $g'^\tau_{\text{ego}}$ by distance $d_{\text{ego-hand}}$ from the head-worn ego camera:
\begin{equation}
    p_{\text{center}} = t'^\tau_{\text{ego}} + d_{\text{ego-hand}} \cdot g'^\tau_{\text{ego}}
\end{equation}
where $g'^\tau_{\text{ego}}$ is orientation vector obtained from the last column of $R'^\tau_{\text{ego}}$.

For each exo camera $i$, we measure visibility of the hand-object interaction region via cosine similarity between its own orientation vector $g'_i$ and the projected vector from the camera position to $p_{\text{center}}$, as illustrated in Fig~\ref{fig:approach_fig}: 
\begin{equation}
    \text{HOI}_i = \text{cos}  (g'_{i}, p_{\text{center}}-t'_{i}).
\end{equation}

This HOI-based ranking addresses \textit{alignment}, but not obstruction --- perfectly centered views may be rendered useless by the ego-camera wearer obstructing the view of the workspace. To address this, we partition $V_{exo}$ into views that are \textit{facing} and \textit{behind} the ego-camera wearer at time $\tau$ using orientation vector cosine similarity in the $XY$ plane:
\begin{align}
    V_{\text{front}} &= \{i \mid i \in \{1,\ldots,N\}, \cos_{XY}(g'^\tau_{\text{ego}}, g'_i) \leq 0\}, \\
    V_{\text{back}} &= \{i \mid i \in \{1,\ldots,N\}, \cos_{XY}(g'^\tau_{\text{ego}}, g'_i) > 0\}.
\end{align}

We order views facing the ego-camera wearer $V_{\text{front}}$ ahead of views located behind the ego-camera wearer $V_{\text{back}}$, then sort views within each set using the HOI-based view-similarity metric:
\begin{align}
    R_\tau = [v_{ego}, \operatorname{sort}_{\downarrow \text{HOI}}(V_{\text{front}}), \operatorname{sort}_{\downarrow \text{HOI}}(V_{\text{back}})],
\end{align}
and cache the per-timestep view ranks for all training takes.

We stress that camera calibration is a fixed one-time overhead cost used for training videos only, and is commonly available in multi-view data collections. Alternatively, ViewBridge can employ auto-calibration methods with a minimal drop in performance (see Supp.~\ref{sec:ablations}). Importantly, camera calibration is never an input to ViewBridge at test time; it takes only arbitrary single-view video.

\section{Downstream keystep tasks and models}
\label{sec:tasks}

We integrate ViewBridge with models for two distinct video understanding tasks: grounding keysteps in complex, untrimmed activity video (Sec.~\ref{sec:grounding}) and recognizing fine-grained keysteps from short, trimmed clips  (Sec.~\ref{sec:keysteps}). We choose these tasks given their sensitivity to occlusions, which makes them a prime testbed for the problem we aim to solve. Keysteps involve subtle motions that can be easily observed in one view and entirely missed in another with even slight occlusions. 

\subsection{Temporal sentence grounding}
\label{sec:grounding}

Temporal sentence grounding (TSG)~\cite{Zhang_2023, flanagan2023learningtemporalsentencegrounding, zeng2020denseregressionnetworkvideo, han2022temporalalignmentnetworkslongterm, mavroudi2023learninggroundinstructionalarticles} localizes temporal spans of sentences~\cite{han2022temporalalignmentnetworkslongterm} or activity keysteps~\cite{mavroudi2023learninggroundinstructionalarticles} in an untrimmed video. We apply TSG to natural exocentric video captured from viewpoints with varying levels of occlusion.

\textbf{Task formulation.}  Given a $T$-second long exo-view video $\mathcal{V}$ and fine-grained activity keystep descriptions $\mathcal{N} = \{n_i\}_{i=1}^N$ that occur during the video clip (e.g. \textit{``Add salt to the noodles in the pot"} for cooking, or \textit{``Fit the new bike inner tube into the bike wheel"} for repairing a bike), we wish to determine the set of temporal intervals during which each keystep is demonstrated in the video $\{[{start}_i,{end}_i]\}_{i=1}^N$.

\begin{figure*}[t]
  \centering
   \includegraphics[width=1.0\linewidth]{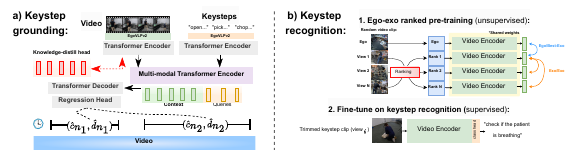}
   \caption{\textbf{Downstream keystep tasks.} \textbf{a)} Our temporal keystep grounding model receives as input an untrimmed video $\mathcal{V}$ and sequence of keysteps $\mathcal{N}$ and regresses the center timestamp $\hat{c}_{n_i}$ and duration $\hat{d}_{n_i}$ for each narration $n_i$. We jointly optimize with our cross-view/cross-temporal knowledge distillation loss (red). \textbf{b)} We pre-train a keystep recognition model on randomly-selected clips from untrimmed videos. We rank the views using our metric and train with our view-contrastive loss that maximizes similarity with the best-exo view.
  }
   \label{fig:downstream_task}
\end{figure*}

\textbf{Approach.} As shown in Fig.~\ref{fig:downstream_task} (a), we extract per-second video features $F = [f_1,\ldots, f_T]$ from $\mathcal{V}$ and keystep features $P = [p_1,\ldots,p_N]$ using video and text embedding models. $F$ and $P$ are fed to modality-specific transformer encoders, and the output video features $F'$ are input to a knowledge-distillation head consisting of an MLP projection layer $l_{proj}$ and our knowledge-distillation objective (Sec.~\ref{sec:k_dist}). The contextualized keystep and video features are concatenated and fed to a multi-modal transformer encoder for cross-modal reasoning. The output keystep features are passed as queries to a transformer decoder with the video features as context, and an MLP head regresses relative center timestamp $0 \leq \hat{c}_{n_i} \leq 1$ and duration $0 \leq \hat{d}_{n_i} \leq 1$ for each keystep.

We train our grounding model with standard L1 loss regression loss applied on the predicted center and duration for each keystep ($\mathcal{L}_{\text{center}}$ and $\mathcal{L}_{\text{dur}}$, respectively) as well as an IoU loss between predicted and ground truth spans of each keystep:
\begin{equation}
    \mathcal{L}_{\text{IoU}}(n_i) = 1 - \frac{|{v}_{n_i} \cap {\hat{v}}_{n_i}|}{|{v}_{n_i} \cup {\hat{v}}_{n_i}|},
\end{equation}
where $v_{n_i} = [s_{n_i}, e_{n_i}]$ denotes the temporal span from $s_{n_i}$ to $e_{n_i}$. The full grounding loss is:
\begin{align} 
    \mathcal{L}_{\text{ground}} = \frac{1}{|N|}\sum\limits_{i=1}^N\lambda_c \mathcal{L}_{\text{center}} +
    \lambda_d \mathcal{L}_{\text{dur}} + \lambda_{\text{iou}} \mathcal{L}_{\text{IoU}},
\end{align}
where $\lambda_c, \lambda_d, \text{and} \lambda_{iou}$ are loss-specific weights. We jointly optimize grounding and knowledge distillation objectives
\begin{equation}
    \mathcal{L}_{\text{combined}} = \mathcal{L}_{\text{ground}} + \lambda_{\text{InfoNCE}}\mathcal{L}_{\text{InfoNCE}}.
\end{equation}

\subsection{Fine-grained keystep recognition}
\label{sec:keysteps}

Keystep recognition is the inverse of grounding: given a trimmed clip from a multi-step procedure, the goal is to name the demonstrated keystep (e.g., ``grease the chain'' in bike repair). It has been studied from both ego \cite{sigurdsson2018charadesegolargescaledatasetpaired, ragusa2020meccanodatasetunderstandinghumanobject, bansal2022viewbestviewprocedure, NEURIPS2023_7a65606f} and exo perspectives \cite{Tang_2021, zhukov2019crosstaskweaklysupervisedlearning, zhou2017automaticlearningproceduresweb, ashutosh2023videominedtaskgraphskeystep, mavroudi2023learninggroundinstructionalarticles}. While Ego-Exo4D~\cite{grauman2024egoexo4dunderstandingskilledhuman} shows that cross-view contextualization can improve an \emph{egocentric} recognition backbone, we instead leverage multi-view training videos to improve \emph{exocentric} recognition, where hands or body pose may be severely occluded.

\textbf{Task formulation.} We treat fine-grained keystep recognition as a trimmed video classification task. We are given a training dataset of paired ego-exo trimmed keystep clips
\begin{equation}
    \mathcal{D} = \{(v^1_{ego},v^1_{exo^{1:N}}, k^1),\ldots,(v^D_{ego},v^D_{exo^{1:N}}, k^D)\}
\end{equation}
where $k^D$ is the keystep class label for sample $D$. At inference, given a single (single-view) exocentric trimmed video clip $v_{exo^i}$, the model must classify the keystep $k$.

\textbf{Approach.} We follow a two-stage approach as introduced in the Ego-Exo4D keystep recognition benchmark~\cite{grauman2024egoexo4dunderstandingskilledhuman}. See Fig.~\ref{fig:downstream_task} (b).  We pre-train a model on random $d$-second length clips, where $d$ is randomly selected from the range of common keystep lengths (Sec.~\ref{sec:imp_details}) from untrimmed videos, followed by fine-tuning on trimmed keystep classification. We first use our view ranking method to determine the best exo-view $v_{best}$ at all seconds in all untrimmed training videos. We use this to pre-train a model $\mathcal{M}$ with an unsupervised contrastive objective

\begin{align}
    \mathcal{L}_{\text{pre-train}} = \text{InfoNCE}(v_{ego}, v_{best}) + \\
    \sum\limits_{exo^{1:N}\setminus v_{best}}\text{InfoNCE}(v_{best}, v_{exo}),
\end{align}
where InfoNCE is the batch-level contrastive loss that treats features from mismatched samples as negatives and features from the same sample as a positive. $\mathcal{L}_{\text{pre-train}}$ jointly forces alignment between ego and best exo-view $v_{best}$, along with exo-exo alignment between $v_{best}$ and each poorer quality view.  We fine-tune the pre-trained model $\mathcal{M}$ on the keystep classification task. Given trimmed exocentric video clip $v_{exo} \in \{v_{exo^{1:N}}\}$ and keystep label $k$, we input $v_{exo}$ to $\mathcal{M}$ and a classification head, and optimize cross-entropy loss between predicted and ground truth keysteps $\hat{k}$ and $k$.

\subsection{Implementation details}
\label{sec:imp_details} 
For TSG, we extract EgoVLPv2~\cite{pramanick2023egovlpv2egocentricvideolanguagepretraining} text keystep and video features at 1 feature per second from a model pre-trained on all views. We split videos into $T$=64 second chunks for input to the grounding model. We allocate a longer portion ($l_{P}=50\%$) of total training epochs $M$ for phase $P$ to allow adaptation to the extreme ego-exo viewpoint-disparity, then evenly divide the remaining $M(1-l_{P})$ epochs among all other phases. See Supp.~Tab.~\ref{tab:curr_ablations} for ablations examining design choices and parameters for the curriculum schedule. The number of phases $P$ is by definition the number of distinct viewpoints; hence ($P$=5 for Ego-Exo4D, $P$=1 for LEMMA, $P$=9 for EPFL). 

We train for $M$=200 epochs (Ego-Exo4D, LEMMA) and $M$=400 (EPFL) given the large number of exo views in EPFL. We set $\lambda_{\text{InfoNCE}}, \lambda_{\text{c}}, \lambda_{\text{d}}, \lambda_{\text{iou}}=$1.0. We compute view rankings every $\tau=1$ second. For keystep recognition, we use TimeSformer~\cite{bertasius2021spacetimeattentionneedvideo} as our video encoder backbone and fine-tune for up to $M$=50 epochs. We pre-train on random clips of length $d \in [6,18]$ seconds based on keystep duration ranges listed in Ego-Exo4D, and $d \in [2,8]$ for LEMMA and EPFL given average action annotation lengths $\sim$3 sec and $\sim$4 sec, respectively.
\section{Experiments}
\label{sec:experiments}

\begin{table*}[t]
  \centering
  \caption{\textbf{Results on two tasks for three datasets.}
  \textbf{(a) Temporal grounding}.
  We report Recall, stratified by the easiest (E), moderate (M), and most difficult (D) views.
  Recall LEMMA provides only one exo viewpoint per take, allowing only D breakdown. \textbf{(b) Keystep recognition.} We report top-1 accuracy (\%).
  We outperform the latest open source SOTA methods from the Ego-Exo4D challenge~\cite{grauman2024egoexo4dunderstandingskilledhuman}. Note that keystep
  recognition cannot be stratified by difficulty; see text. ViewBridge always uses a
  single, uncalibrated exo viewpoint at inference. Higher is better for all metrics.
  }
  \label{tab:model_performance_grounding}

  \vspace{1em}

  \begin{minipage}[t]{0.48\linewidth}
    \centering
    (a) Temporal grounding
    \normalsize
    \setlength{\tabcolsep}{1.5pt}
    \renewcommand{\arraystretch}{1.05}
    \makebox[\linewidth][c]{%
      \begin{tabular}{@{}l ccccccc@{}}
        \toprule
        & \multicolumn{1}{c}{LEMMA}
        & \multicolumn{3}{c}{Ego-Exo4D}
        & \multicolumn{3}{c}{EPFL} \\
        \cmidrule(lr){2-2} \cmidrule(lr){3-5} \cmidrule(l){6-8}
        Method
        & D
        & E & M & D
        & E
        & M
        & D \\
        \midrule
        CliMeR~\cite{flanagan2023learningtemporalsentencegrounding}
        & .09
        & .02 & .02 & .02
        & .17 & .14 & .12 \\

        VI Encoder~\cite{grauman2024egoexo4dunderstandingskilledhuman}
        & .15
        & .18 & .18 & .18
        & .34 & .33 & .26 \\

        EgoVLPv2~\cite{pramanick2023egovlpv2egocentricvideolanguagepretraining}
        & .16
        & \textbf{.27} & .27 & .20
        & .34
        & .32
        & .27 \\

        LangView~\cite{majumder2025viewpointshowsbestlanguage}
        & --
        & .24 & .21 & .20
        & .31 
        & .28 
        & .25 \\

        \rowcolor{gray!20}
        \textbf{ViewBridge}
        & \textbf{.18}
        & .26 & \textbf{.28} & \textbf{.28}
        & \textbf{.36}
        & \textbf{.35}
        & \textbf{.31} \\
        \bottomrule
      \end{tabular}%
    }
  \end{minipage}
  \hfill
  \begin{minipage}[t]{0.48\linewidth}
    \centering
    (b) Keystep recognition
    \normalsize
    \setlength{\tabcolsep}{1.5pt}
    \renewcommand{\arraystretch}{1.1}
    \makebox[\linewidth][c]{%
      \begin{tabular}{@{}l c c c c@{}}
        \toprule
        Method & Train & LEMMA & Ego-Exo4D & EPFL \\
        \midrule
        TimeSFormer~\cite{bertasius2021spacetimeattentionneedvideo}
        & Exo & 16.02 & 19.74 & 15.26 \\

        TimeSFormer
        & Ego,Exo & 23.69 & 20.11 & 17.04 \\

        VI Encoder~\cite{grauman2024egoexo4dunderstandingskilledhuman}
        & Exo & 17.16 & 20.44 & 17.83 \\

        VI Encoder
        & Ego,Exo & -- & 23.06 & 18.21 \\

        LangView~\cite{majumder2025viewpointshowsbestlanguage}
        & Ego,Exo & -- & 22.48 & 17.46 \\

        \rowcolor{gray!20}
        \textbf{ViewBridge}
        & \textbf{Ego,Exo}
        & \textbf{27.86}
        & \textbf{24.07}
        & \textbf{19.24} \\
        \bottomrule
      \end{tabular}%
    }
  \end{minipage}
\end{table*}

\subsection{Datasets} 
Departing from the highly edited YouTube video and single-camera setups typically used in large-scale benchmarks~\cite{miech2019howto100mlearningtextvideoembedding, NEURIPS2023_9d58d85b, caba2015activitynet}, we validate ViewBridge in natural, multi-view procedural activity settings---long, untrimmed videos captured in cluttered environments from diverse viewpoints that induce significant occlusions.

(1) \textbf{Ego-Exo4D~\cite{grauman2024egoexo4dunderstandingskilledhuman}} A large-scale, diverse, multi-view dataset consisting of simultaneously captured egocentric and exocentric videos across 43 diverse human activities. Each take consists of an egocentric camera worn by the activity demonstrator, along with synchronous exocentric video from 4-6 stationary cameras at arbitrary positions in the scene (not fixed across takes or environments), capturing diverse viewpoints. Ego-Exo4D includes 664 unique fine-grained keysteps across diverse tasks (e.g., cooking, bike repair, Covid testing). We train/validate on the official keystep recognition train split, then---importantly---evaluate on all \emph{exo} clips from the validation split (vs.~the ego clips used in~\cite{grauman2024egoexo4dunderstandingskilledhuman}), thereby increasing the occlusion level and difficulty.

(2) \textbf{LEMMA~\cite{jia2020lemmamultiviewdatasetlearning}} A multi-view video dataset for goal-directed activity capturing 324 diverse household activities across 14 environments, with one ego camera and one exo camera per scene.\footnote{LEMMA has only one exocentric view per take. Thus, experiments that require multiple exo candidates---E/M/D view-rank stratification and learned best-view selection among exo views---are not meaningful. We report a single LEMMA exo result and mark non-applicable or degenerate entries with `--'.} LEMMA features 11,781 hand-object interactions (HOIs), 641 compositional action labels, and 4.6 million frames (RGB + depth), with 0.9 million annotated frames. We focus on the single-agent single-task split, and use the given (single) exo view label as the `Difficult' view rank.

(3) \textbf{EPFL-Smart-Kitchen-30~\cite{bonnetto2025epflsmartkitchen30denselyannotatedcooking}} A densely annotated multi-view cooking dataset captured in an instrumented kitchen, containing 29.7 hours of synchronized recordings from 16 subjects across 49 sessions, captured by one egocentric HoloLens~2 headset and 9 calibrated static RGB-D exocentric cameras. The dataset provides 763 fine-grained action classes composed from 33 verbs and 79 nouns, totaling 60,189 action segments. We report results on the \textit{exo-view} video clips in the action recognition benchmark test split, rather than the ego-view, to measure performance under viewpoint variation and challenging occlusion.

\subsection{Baselines} 
We compare against state-of-the-art (SOTA) view-invariant representation learning and grounding methods, with emphasis on methods that exploit ego-exo synchronized video and grounding methods that learn from video recorded from a single viewpoint (as opposed to edited YouTube video).
\begin{itemize}
    \renewcommand\labelitemi{--}
    \item \textbf{CliMer}~\cite{flanagan2023learningtemporalsentencegrounding}: a TSG method that predicts boundaries between short video clips belonging to different narrations. We train CliMer on ego+exo video and test on exo views.
    \item \textbf{VI Encoder}~\cite{grauman2024egoexo4dunderstandingskilledhuman}: the SOTA Ego-Exo4D keystep recognition baseline, a TimeSformer~\cite{bertasius2021spacetimeattentionneedvideo} model trained with a clip-level Ego-Exo contrastive loss. We fine-tune VI Encoder for keystep recognition, and also train a TSG model using VI Encoder as the video encoder backbone.
    \item \textbf{TimeSformer}~\cite{bertasius2021spacetimeattentionneedvideo}: For action recognition, we evaluate TimeSformer models pre-trained on Kinetics-600~\cite{carreira2018shortnotekinetics600} and finetuned on 1) exo-view keystep video clips and 2) ego+exo-view keysteps clips.
    \item \textbf{EgoVLPv2}~\cite{pramanick2023egovlpv2egocentricvideolanguagepretraining}: SOTA video-text embedding model trained via contrastive clip-keystep loss. We use EgoVLPv2 trained on all views (ego+exo) to generate weakly-view invariant features (aligned  by keystep text).
    \item \textbf{LangView}~\cite{majumder2025viewpointshowsbestlanguage}: a learned best-view selection method; we use LangView to identify the best view at regular temporal intervals, and train a grounding model with the cross-view positive from our distillation objective. See~Supp.~\ref{sec:exp_imp_details} for implementation details.
\end{itemize}

\begin{figure*}
  \centering
   \includegraphics[width=1.0\linewidth]{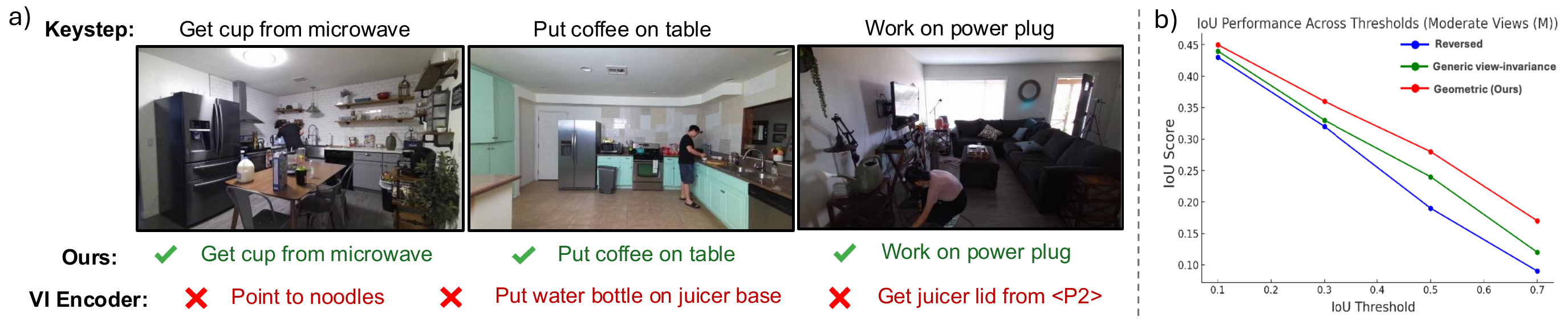}
   \caption{\textbf{a) Keystep recognition on LEMMA.} ViewBridge performs strongly across diverse tasks on video that exhibit natural occlusions of the interacted objects or actions. Best in zoom. \textbf{b) Evaluation of different ranking strategies.} Training a TSG model with our view ranking (red) vs. reversed or no rankings yields minimal drop in performance across IoU thresholds on moderate difficulty views. }
   \label{fig:ksr_lemma}
\end{figure*}

\vspace{0.1in}
We evaluate CliMer, EgoVLPv2, and VI Encoder on the TSG task as they either 1) learn from \textit{single-view} video or 2) are SOTA view-invariant encoders. For keystep recognition, we re-train and evaluate for exo-view inference the top-performing open-source models on the Ego-Exo4D keystep recognition benchmark, VI Encoder and TimeSformer pre-trained on Kinetics-600. We evaluate LangView on both tasks for comparison against a SOTA learned best-view ranking method.

\paragraph{Metrics.} We report standard metrics for each task: Recall@$1$ with Intersection-over-Union (IoU) $\geq \theta$ for TSG~\cite{flanagan2023learningtemporalsentencegrounding, barrios2023localizingmomentslongvideo}, and top-1 accuracy (\%) for keystep recognition. We set $\theta$ according to the temporal scale of
each dataset. LEMMA and EPFL actions are extremely short ($\sim$3s, 4\% of the video window), so even a 1 second temporal offset---well within typical annotation variability---drastically reduces IoU.
Ego-Exo4D keysteps are substantially longer (6–18s, or 10–28\% of the video). We thus report $\theta=0.1$ on LEMMA and EPFL and $\theta=0.5$ on Ego-Exo4D to normalize evaluation difficulty across datasets with markedly different temporal scales. See~Supp.~\ref{sec:full_gnd_results} for full results across multiple thresholds $\theta$.

\subsection{Temporal sentence grounding}
Table~\ref{tab:model_performance_grounding} (a) shows the results for TSG. We break down results as a function of input view difficulty, ranging from easy (E) to moderate (M) to difficult (D), according to camera ranks. Recall that our method ought to have the greatest advantage precisely on the most difficult viewpoints that endure significant occlusions.

We outperform SOTA ego-exo view-invariant representation learning method, VI Encoder, as well a SOTA egocentric grounding model, CliMer. We observe poor performance with CliMer, suggesting that their contrastive stitching strategy breaks down in non-egocentric settings, where exo clips from very different keysteps can still share significant visual overlap. This underscores the complexity of learning temporally discriminative features from exo views, particularly from distant or occluded viewpoints where visual content is largely stationary. 

Our result compared to VI Encoder (row 2) validates that the standard ego-exo contrastive training is insufficient. EgoVLPv2 performs well on easy views, but drops by 26\% relative to easy on difficult Ego-Exo4D views, indicating that generic video-text alignment does not by itself resolve viewpoint-induced occlusion either. In contrast, ViewBridge \textit{improves} on moderate/difficult views on Ego-Exo4D (+8\% relative to easy views), consistent with our goal of enriching lower-quality views using better-ranked synchronous views. On EPFL, where viewpoint diversity is larger, ViewBridge drops only 14\% relative from easy to difficult views, compared to roughly 19--29\% for baselines.

Overall, ViewBridge is substantially more robust to viewpoint difficulty. On Ego-Exo4D, performance improves from easy to difficult views, while on EPFL, where viewpoint diversity is larger, its relative difficult-view degradation is smaller than the baselines. This supports our central hypothesis: lower-ranked, occluded views benefit most when training explicitly distills information from better-ranked synchronous views. \textbf{Ablations} in Supp.~Sec.~\ref{sec:ablations} further confirm that the curriculum is necessary for realizing the full benefit of ranked distillation.

\subsection{Keystep recognition} 
Next we evaluate on the single-view keystep recognition task.  Table~\ref{tab:model_performance_grounding} (b) reports standard top-1 accuracy ~\cite{grauman2024egoexo4dunderstandingskilledhuman, bertasius2021spacetimeattentionneedvideo, carreira2018shortnotekinetics600} across all \textit{exo} views.\footnote{Unlike TSG, breaking down keystep results by view rank is not applicable -- significant ego motion during keystep intervals means multiple distinct view ranks occupy any given keystep temporal segment.} We significantly outperform the latest open-source leader method on the Ego-Exo4D fine-grained keystep recognition benchmark (VI Encoder). This demonstrates that our ranking-based view-contrastive loss improves upon generic view-invariance when applicable in multi-view setups. Notably, LangView performs similarly poorly to the generic VI Encoder across both tasks, suggesting that despite directly distilling the global best-view at each moment, neglecting to address the viewpoint disparity limits performance.

Fig.~\ref{fig:ksr_lemma} visualizes keystep recognition results on LEMMA against VI Encoder; ViewBridge recognizes keysteps in the presence of natural self- and object-occlusions across diverse scenarios such as cooking and home repair. \textbf{Ablations} in Supp.~Sec.~\ref{sec:ablations} also show that ViewBridge's success is not dependent on ground-truth calibration: rankings computed from automatically estimated camera poses still improve over LangView as well as the no-distillation and generic ego-exo distillation settings, while remaining close to the calibrated setting performance (see~Supp.~Tab.~\ref{tab:supp_ablations})

\subsection{Impact of ranking strategy}
Finally, we assess the value of our view ranking strategy.  Fig.~\ref{fig:ksr_lemma} (b) evaluates ranking strategies on moderate-difficulty Ego-Exo4D views. We evaluate our view ranking (red) against generic view-invariance (green) which selects a target view at random, and against our view rankings reversed in order (blue) on moderate difficulty views. Reversing our ranking actively hurts performance (below generic view-invariance), indicating that our view rankings correctly assess view quality. We provide full results on all views in Supp. Together with our improvement over SOTA learned best-view selection methods (Tab.~\ref{tab:model_performance_grounding}, LangView), this convincingly validates the robustness of our ranking design. Additional results providing the ablations discussed above, analysis of our curriculum design, an experiment that omits the ego view during training, and a video with examples and failure cases are provided in~Supp. below.
\section{Conclusion}
\label{sec:conclusion}
We present \textbf{ViewBridge} for learning video representations from multi‑view footage with large viewpoint gaps. We define a metric that ranks views by mutual visibility of the acted region alongside a knowledge‑distillation objective that enriches low‑quality views using visually richer ones. To handle extreme viewpoint shifts, we use a curriculum that gradually distills between incrementally different views during training, enabling smoother adaptation. Across two tasks and three datasets, ViewBridge yields clear gains, paving the way for activity understanding in in-the-wild, unedited video with significant viewpoint variation and occlusion. Future work could move beyond synchronized captures toward learning from asynchronous videos, where models must automatically discover which views overlap in time and which best observe the activity.

{
    \small
    \bibliographystyle{ieeenat_fullname}
    \bibliography{main}
}
\clearpage
\section{Appendix}
\label{sec:table_of_contents}
\begin{enumerate}
\item \textbf{Expanded implementation details (Sec.~\ref{sec:exp_imp_details}) ---} We provide further implementation details regarding hyperparameters, baseline evaluations, and training/evaluation settings.

\item \textbf{Full keystep grounding results (Sec.~\ref{sec:full_gnd_results}) ---} We report the full version of Table~\ref{tab:model_performance_grounding} across multiple IoU thresholds $\theta$, as mentioned in Sec.~\ref{sec:experiments} (Temporal Keystep Grounding) of the main paper.

\item \textbf{Ablations of ViewBridge and evaluation with uncalibrated cameras (Sec.~\ref{sec:ablations}) ---} We provide ablations of each component of ViewBridge, as well as TSG results using rankings obtained entirely calibration-free.

\item \textbf{Keystep grounding results stratified by keystep name and task (Sec.~\ref{sec:keystep_strat}) ---} We provide an analysis of our model's performance relative to EgoVLPv2 (strongest baseline) within each unique keystep name as well as within each high-level activity.
\item \textbf{Feature similarity with ego feature vs. EgoVLPv2 (Sec.~\ref{sec:ego_feat_align}) ---} We provide an analysis demonstrating close alignment between our learned features from any source view and the corresponding ego video features at each moment as verification of effective distillation between target and source views.
\item \textbf{Results on keystep grounding in seen and unseen environments (Sec.~\ref{sec:seen_vs_unseen}) ---} We stratify our test set by videos from environments observed during training (test-seen) and from environments unseen during training (test-unseen) to evaluate robustness of our approach to novel scenes.
\item \textbf{Ablations of camera ranking algorithm/use. (Sec.~\ref{sec:cam_rank_exper}) ---} We train a model with several variations of our camera ranking to quantitatively validate its utility vs. selecting a random distillation target, as well as to confirm that our particular camera ranking is effective.
\item \textbf{Ablation of our curriculum design. (Tab.~\ref{tab:curr_ablations}) ---} We evaluate ablations of our curriculum without any ego distillation, and with varying lengths of ego-distill phases $l_{P}$.
\item \textbf{t-SNE of learned video features. (Sec.~\ref{sec:tsne}) ---} We provide a t-SNE visualization of our learned video features and their similarity with the best-view features at each instant.
\item \textbf{Supplemental video.} We include a short video with examples of our camera ranking across diverse scenarios, as well as qualitative keystep grounding and recognition examples compared against the SOTA models for reference, as well as failure cases (see attached video).
\end{enumerate}

\subsection{Expanded implementation details}
\label{sec:exp_imp_details}

We use an Ego-Exo4D trained checkpoint provided by the authors for LangView inference on Ego-Exo4D and EPFL. As EPFL has nine exo views, we select the top five views in EPFL which have highest average view rank under our own ViewBridge rankings across all takes and pass these as input to LangView. For Ego-Exo4D keystep recognition, we follow the benchmark design in~\cite{grauman2024egoexo4dunderstandingskilledhuman} and restrict training and evaluation to the 278 unique keystep labels with $>$20 instances in the dataset. Baselines are implemented using the author's codebases. We follow the training procedure/hyperparameters stated in CliMer~\cite{flanagan2023learningtemporalsentencegrounding}; for VI Encoder, TimeSformer, and EgoVLPv2, we use the training procedure/hyper-parameters stated in the Ego-Exo4D fine-grained keystep recognition benchmark (Ego-Exo4D~\cite{grauman2024egoexo4dunderstandingskilledhuman}, Sec. 5.2.1). All models are trained on 8 GH200 NVIDIA GPUs.  

\subsection{Complete keystep grounding results}
\label{sec:full_gnd_results}
\begin{figure*}
  \centering    \includegraphics[width=0.99\linewidth]{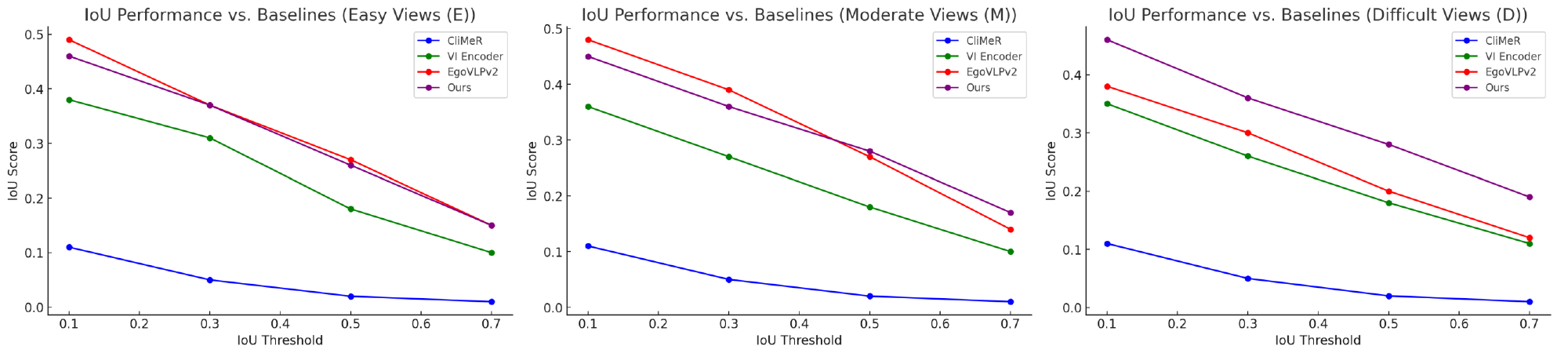}

   \caption{\textbf{Temporal keystep grounding stratified by view quality on Ego-Exo4D.} We visualize $\text{Recall}@1$ with IoU at threshold $\theta$ on easy (E), moderate (M), and difficult (D) views, where quality is judged by the amount of activity occlusion. Our method consistently outperforms baselines at high IoU thresholds ($
   \theta=0.5,0.7$), particularly in difficult views (D), where occlusion is severe. While baselines such as EgoVLPv2 perform competitively at low thresholds, they suffer significant performance drops at higher IoU levels. Weakly supervised approaches like CliMeR struggle across all settings. These results underscore the robustness of our approach in handling challenging viewpoints.}
   \label{fig:vis_tmp_gnd}
\end{figure*}

We report Table~\ref{tab:model_performance_grounding} from Experiments~\ref{sec:experiments} in the main text with a wider set of IoU thresholds in Table~\ref{tab:full_grounding} and in Fig.~\ref{fig:vis_tmp_gnd}. As observed in Table~\ref{tab:model_performance_grounding}, we outperform existing methods on the most challenging views with severe occlusion (D) across several IoU thresholds $\theta$, and particularly outperform at high IoU thresholds on \textit{all} views, including on easy views (E) and views with moderate view difficulty (M).

\subsection{ViewBridge ablations and results with calibration-free rankings.}
\label{sec:ablations}
Table~\ref{tab:supp_ablations} reports ablations. Training without view distillation (row 1) leads to poor performance. Using distillation based on our view rankings (rows 3-6) outperforms generic ego-exo view distillation (row 2). While training with our full knowledge distillation loss (row 5) performs well, the addition of the curriculum (rows 6 and 7) significantly boosts performance, demonstrating the benefit of curriculum learning \textit{alongside} knowledge distillation. Finally, ViewBridge is fairly robust to training with cameras self-localized with Reloc3r~\cite{dong2025reloc3rlargescaletrainingrelative} (compare last two rows). For LEMMA, view-rank and calibration ablations do not apply since it has only a single exo view.

\begin{table*}[t]
    \centering
    \normalsize
    \caption{\textbf{Ablations on TSG.} We ablate the components of ViewBridge on temporal keystep grounding and report IoU$\geq \theta$ for Ego-Exo4D ($\theta=0.5$), LEMMA and EPFL ($\theta=0.1$) below. Curriculum learning significantly boosts performance when used with ranked knowledge distillation (last row), outperforming generic view-invariant features (row 1) or direct ego-exo distillation (row 2). The row marked View-ranked* uses rankings computed from automatically estimated camera poses rather than ground-truth calibration; it still improves over no distillation and generic ego-exo distillation, showing that ViewBridge can benefit from noisy, estimated rankings.}
    \small
    \setlength{\tabcolsep}{3pt}
\begin{tabular}{lllll|c|c|c}
Distill & \shortstack{View\\-ranked} & \shortstack{Cross\\-view} & \shortstack{Same\\-view} & Curr. & \multicolumn{1}{c}{Ego-Exo4D} & \multicolumn{1}{c}{LEMMA} & \multicolumn{1}{c}{EPFL} \\ 
\hline
\textcolor{red}{\ding{55}} & \textcolor{red}{\ding{55}} & \textcolor{red}{\ding{55}} & \textcolor{red}{\ding{55}} & \textcolor{red}{\ding{55}} & .20 & .16 & .27\\
\textcolor{green}{\ding{51}} & \textcolor{red}{\ding{55}} & \textcolor{green}{\ding{51}} & \textcolor{red}{\ding{55}} & \textcolor{red}{\ding{55}} & .18 & .15 & .25\\
\textcolor{green}{\ding{51}} & \textcolor{green}{\ding{51}} & \textcolor{green}{\ding{51}} & \textcolor{red}{\ding{55}} & \textcolor{red}{\ding{55}} & .26 & -- & .26\\
\textcolor{green}{\ding{51}} & \textcolor{green}{\ding{51}} & \textcolor{red}{\ding{55}} & \textcolor{green}{\ding{51}} & \textcolor{red}{\ding{55}} & .25 & -- & .27\\
\textcolor{green}{\ding{51}} & \textcolor{green}{\ding{51}} & \textcolor{green}{\ding{51}} & \textcolor{green}{\ding{51}} & \textcolor{red}{\ding{55}} & .24 & -- & .27\\ 
\textcolor{green}{\ding{51}}  & \textcolor{green}{\ding{51}}* & \textcolor{green}{\ding{51}} & \textcolor{green}{\ding{51}} & \textcolor{green}{\ding{51}} & .26 & .18 & .29\\
\rowcolor{gray!20} \textcolor{green}{\ding{51}}  & \textcolor{green}{\ding{51}} & \textcolor{green}{\ding{51}} & \textcolor{green}{\ding{51}} & \textcolor{green}{\ding{51}} & \textbf{.28} & \textbf{.18} & \textbf{.31}\\
\bottomrule
\end{tabular}
\label{tab:supp_ablations}
\end{table*}

\subsection{Results stratified by keystep name and activity on Ego-Exo4D}
\label{sec:keystep_strat}
\begin{figure*}
  \centering
   \includegraphics[width=0.95\linewidth]{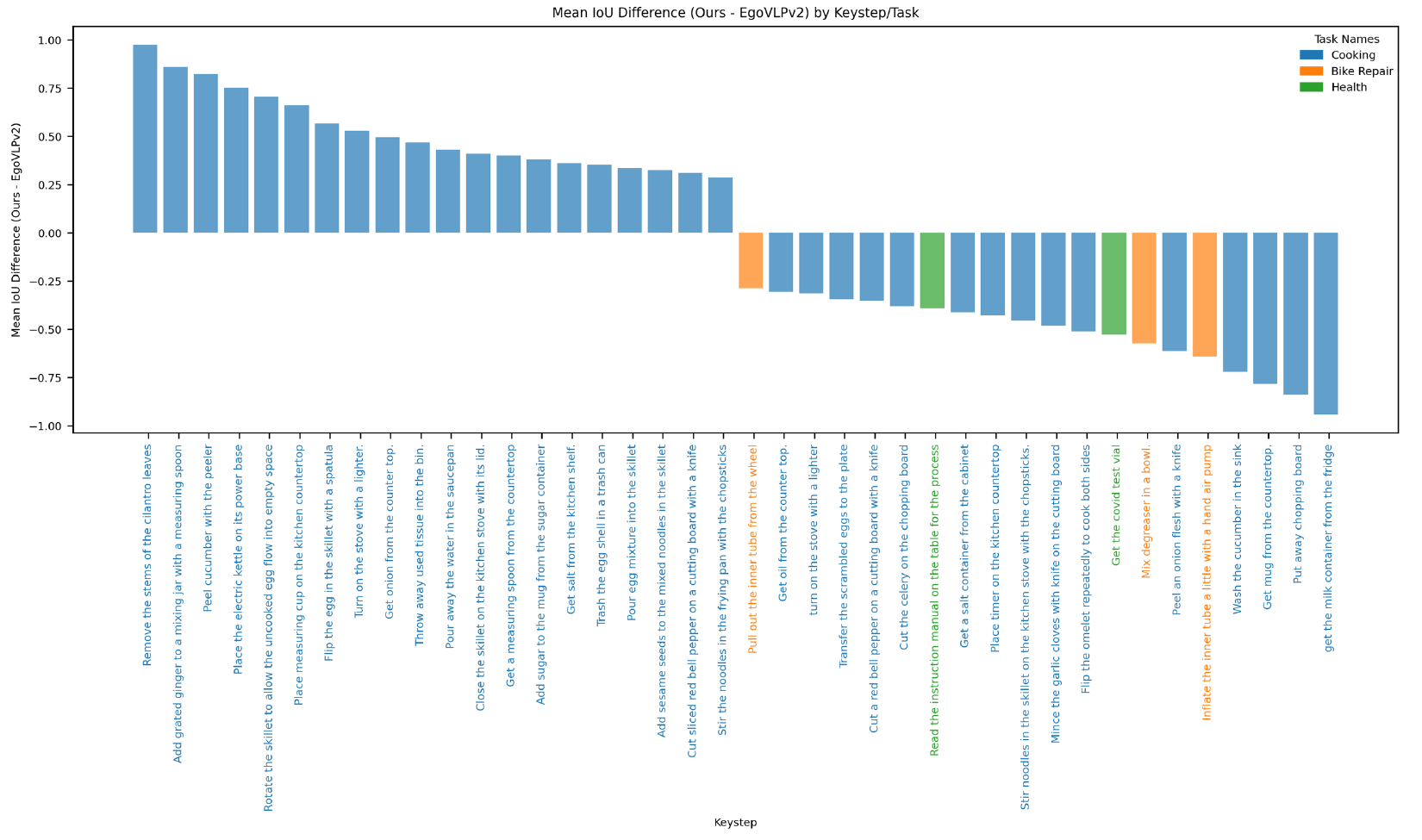}

   \caption{\textbf{Mean IoU difference (ViewBridge - EgoVLPv2) by keystep name and task.} We compute mean IoU across all instances and views of each unique keystep in the test set -- for both our model and the EgoVLPv2-trained grounding model. We display signed mean IoU difference between ours and EgoVLPv2 for the top-20 keysteps (left half) and bottom-20 keysteps (right half) that have largest mean IoU difference. We outperform EgoVLPv2 on keystep names that require an unobstructed view of fine-grained actions, despite being associated with cooking activities (blue) which exhibit widest viewpoint diversity in our dataset.
   }
   \label{fig:keystep_strat_fig}
\end{figure*}

We compute mean IoU across all occurrences of each unique keystep name in the EgoExo4D test set, and compute the signed difference between our model and the EgoVLPv2-trained model mean IoU for each keystep name. Figure~\ref{fig:keystep_strat_fig} visualizes this for top-20 keysteps where our model best outperforms EgoVLPv2 (left) and the bottom-20 keysteps where EgoVLPv2 best outperforms ours. We observe that our model strongly outperforms EgoVLPv2 on narrations from cooking scenarios -- as shown by the solid blue bars/keysteps on the left hand side of the plot. Indeed, cooking scenarios display the largest variability in viewpoint and workspace occlusion compared to other high-level activity categories in our dataset (bike repair, CPR, taking a covid test), demonstrating that our method performs strongest in these activity/capture settings that most resemble in-the-wild video. We observe a natural divide within cooking-related keysteps as well; EgoVLPv2 best outperforms our model on cooking keysteps that involve significant body movement easily visible in all views (\textit{"get the milk container from the fridge"}, \textit{"put away chopping board"}, \textit{"get mug from the countertop"}, etc.), whereas our model outperforms EgoVLPv2 strongly on more subtle keysteps that require privileged information from optimally placed views: (\textit{"remove the stems of the cilantro leaves"}, \textit{"add grated ginger to a mixing jar..."}, \textit{"peel cucumber with the peeler"}, etc.).

\subsection{Ego-feature alignment}
\label{sec:ego_feat_align}
To measure the effectiveness of our knowledge distillation objective, we evaluate how well our learned features discriminate between features that share the same view but different action (same-view negative), features that share the same action from the most occluded viewpoint (cross-view negative), and the synchronized ego-view feature -- on the test set. We report our results with both EgoVLPv2 features and our learned features in Table~\ref{tab:ego_sim}. 'Avg neg cos sim' reports cosine similarity between source-view feature and the negative features (cross-view negative, same-view negative), and 'InfoNCE loss' computes cosine similarity with the ego-view feature \textit{relative} to all other negative/positive features (see Sec.~\ref{sec:k_dist} for our modified InfoNCE metric). We report results stratified by view difficulty (easy view (E), moderate view (M), and difficult view (D)).

We observe significant reduction in InfoNCE loss from EgoVLPv2 features to our learned features on the test set videos, indicating that our model successfully learns temporally discriminative, action-centric features that are not only closely aligned with the visually rich ego-view feature, but distinct from superficially similar features that come from 1) the same view and 2) the same (synchronous) action, but a severely occluded viewpoint.

\begin{table}[t]
    \centering
    \caption{\textbf{Ego/source-view alignment on features learned by our model vs EgoVLPv2.} We report results stratified by source view quality (easy views (E), moderate views (M), difficult views (D)).}
    \setlength{\tabcolsep}{2pt}
    \begin{tabular}{lccc|ccc}
        & \multicolumn{3}{c|}{Avg neg cos sim ($\downarrow$)} & \multicolumn{3}{c}{InfoNCE loss ($\downarrow$)} \\
        Features & E & M & D & E & M & D \\
        \hline
        EgoVLPv2~\citep{pramanick2023egovlpv2egocentricvideolanguagepretraining} & .44 & .43 & .41 & 4.2 & 3.5 & 3.1 \\
        \rowcolor{gray!20} \textbf{ViewBridge} & \textbf{\textminus.03} & \textbf{\textminus.02} & \textbf{\textminus.03} & \textbf{.80} & \textbf{.95} & \textbf{1.1} \\
        \bottomrule
    \end{tabular}
    
\label{tab:ego_sim}
\end{table}

\begin{table*}[t]
\centering
\caption{\textbf{Temporal keystep grounding stratified by view quality across IoU thresholds.} We report $\text{Recall}@1$ at two IoU thresholds for each dataset on easy (E), moderate (M), and difficult (D) views. We use dataset-specific IoU thresholds because the temporal scale of annotated actions differs substantially: Ego-Exo4D keysteps are longer, so stricter thresholds such as IoU$\geq$0.5 and IoU$\geq$0.7 are meaningful, whereas EPFL actions are much shorter, making IoU$\geq$0.1 and IoU$\geq$0.3 more comparable in localization difficulty. Results are rounded to the nearest $10^{-2}$.}
\normalsize
\setlength{\tabcolsep}{3pt}
\begin{tabular}{lcccc}
\toprule
 & \multicolumn{2}{c}{Ego-Exo4D} & \multicolumn{2}{c}{EPFL} \\
\cmidrule(lr){2-3} \cmidrule(lr){4-5}
 & IoU$\geq$0.5 & IoU$\geq$0.7 
 & IoU$\geq$0.1 & IoU$\geq$0.3 \\ 
Model & E/M/D & E/M/D & E/M/D & E/M/D \\ 
\midrule
CliMeR~\cite{flanagan2023learningtemporalsentencegrounding} 
& .02/.02/.02 & .01/.01/.01
& .17/.14/.12 & .11/.08/.08 \\

VI Encoder~\citep{pramanick2023egovlpv2egocentricvideolanguagepretraining} 
& .18/.18/.18 & .10/.10/.11
& .34/.33/.26 & .23/.19/.15 \\

EgoVLPv2~\citep{pramanick2023egovlpv2egocentricvideolanguagepretraining} 
& \textbf{.27}/.27/.20 & .15/.14/.12
& .34/.32/.27 & .24/.19/.16 \\

LangView~\citep{majumder2025viewpointshowsbestlanguage} 
& .24/.21/.20 & .12/.12/.11
& .31/.28/.25 & .23/.18/.15 \\

\rowcolor{gray!20} 
\textbf{ViewBridge} 
& .26/\textbf{.28}/\textbf{.28} & \textbf{.15}/\textbf{.17}/\textbf{.19}
& \textbf{.36}/\textbf{.35}/\textbf{.31} & \textbf{.26}/\textbf{.21}/\textbf{.18} \\ 
\bottomrule
\end{tabular}
\label{tab:full_grounding}
\end{table*}

\subsection{Evaluation on seen vs. unseen environments in Ego-Exo4D}
\label{sec:seen_vs_unseen}
We stratify our test set into videos that are recorded in physical environments which were observed during training (test-seen), and videos recorded in five "unseen" environments that were unobserved during training (test-unseen). We focus our test-unseen evaluation on the most viewpoint diverse activity domains -- cooking and bike repair -- given the rigid uniformity of the camera setups and clinical environment setting in health activities (covid testing, CPR). We report our results on both test-seen and test-unseen splits against our strongest baseline (EgoVLPv2) in Table~\ref{tab:seen_vs_unseen}, aggregated across all views. 

Across all IoU thresholds, we outperform EgoVLPv2 on both new videos from seen environments (test-seen), but also by significant margins on new, viewpoint and occlusion-diverse videos of complex cooking and bike repair tasks in settings that were \textit{unobserved} during training --- demonstrating our model's robustness to video from arbitrary viewpoints in unseen environments.

\subsection{Camera ranking ablations on Ego-Exo4D}
\label{sec:cam_rank_exper}
\begin{figure*}[t]
  \centering
  \includegraphics[width=0.99\linewidth]{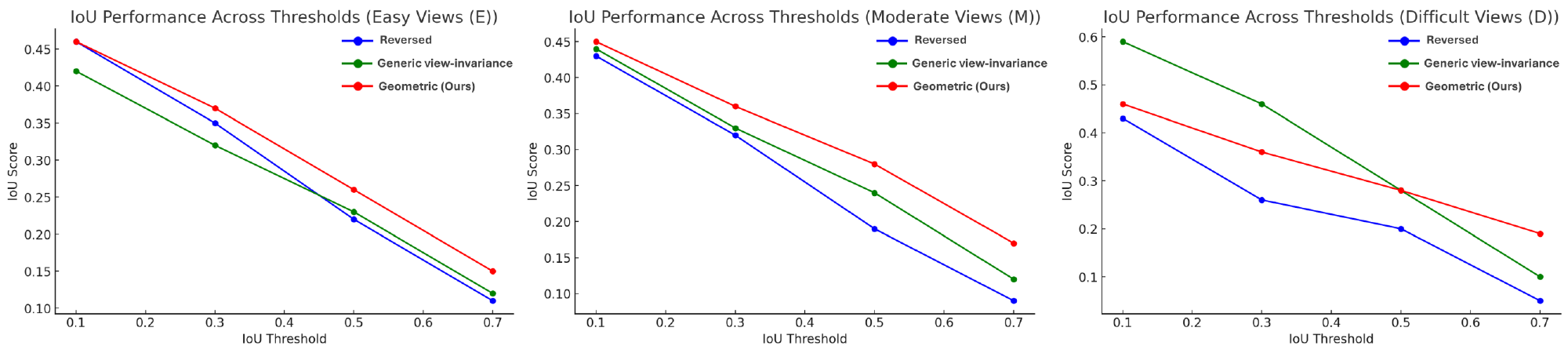}
   \caption{\textbf{Keystep grounding on Ego-Exo4D with various camera ranking strategies.} We report results on easy views (E), moderate views (M) and difficult views (D), where quality is judged by amount of activity occlusion.  $\text{Recall}@K$ with IoU at threshold $\theta$ is reported as IoU@$\theta$. Reversing our geometry-based ranking ("Reversed") underperforms a "Random" method of view selection across all views and IoU thresholds, confirming the validity of our ranking. While generic view-invariance (selecting "Random" pairs for view distillation) achieves good performance on Difficult views at a low IoU threshold, it rapidly deteriorates at high IoU thresholds. In contrast, our geometry-based ranking outperforms generic view-invariance ("random") and our reversed rankings on both Easy and Moderate views at all IoU thresholds (left, middle). On Difficult views, we undergo minimal performance drop at higher IoU thresholds compared to the baseline ranking methods.}
   \label{fig:camera_rank_ablation}
\end{figure*}

We ablate the use of our ranking by training a model that selects another view as the cross-view positive at random during training ("Generic view-invariance"), and validate our particular choice of ranking by training a model that uses the \textit{reverse} of our camera rankings at each second (``Reversed") -- e.g. best-exo becomes worst-exo, and vice versa. We show our results in Fig.~\ref{fig:camera_rank_ablation}, compared against our results with our original geometry-based camera rankings (``Geometric"). We observe that reversing our rankings produces a significant drop in performance below the method that \textit{randomly} selects cross-view distillation positives, confirming that our camera ranking is indeed meaningful.While generic-view invariance performs competitively at low IoU thresholds, our ranking ("Geometric") quickly outperforms this generic view-invariant baseline at higher IoU thresholds, across views of varying difficulty. This further supports our hypothesis that generic 'view-invariance' is insufficient to address the viewpoint and occlusion diversity present in these challenging activities.

\newpage

\subsection{Visualization of learned video features}
\label{sec:tsne}
\begin{figure*}
  \centering
   \includegraphics[width=0.5\linewidth]{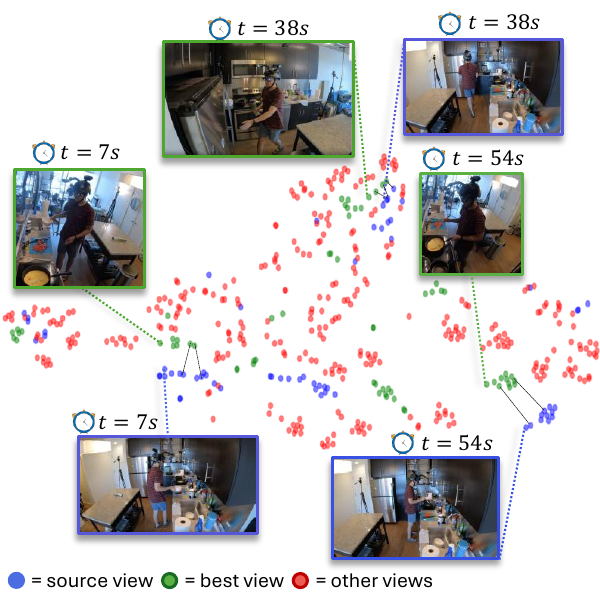}
   \caption{\textbf{t-SNE of learned video features.} We visualize video features learned by our grounding model's knowledge distill head (blue), best-view video features (green), and features from other synchronized views (red) on an input chunk of video. Our model closely aligns source view features with the best-view features throughout the video, \textit{despite} the time-varying nature of the 'best view'.}
   \label{fig:vis_fig}
\end{figure*}

Fig.~\ref{fig:vis_fig} provides a t-SNE visualization of video features produced by our model's knowledge distill head (blue), `best-view' video features (green), and features from other synchronized views (red) on an input video from Ego-Exo4D. Our model consistently outputs video features that are aligned with the best-view, despite this `best-view' shifting among the camera viewpoints from timestep to timestep---unlike in generic view-invariance.

\begin{table*}
\centering
\caption{\textbf{Evaluation on test-seen and test-unseen splits.} We split our test set into videos from scenarios that have been observed during training (test-seen), and videos from five scenarios that were unobserved during training (test-unseen) consisting of cooking and bike repair videos. We report IoU metrics at all thresholds $\theta$. We strongly outperform EgoVLP on both videos from seen and unseen environments --- demonstrating our model's capability to generalize to unseen environments.}
\small
\setlength{\tabcolsep}{2pt}
\begin{tabular}{lcccc|cccc|cccc}
 & \multicolumn{4}{c}{Test-seen} & \multicolumn{8}{c}{Test-unseen} \\ 
  \cmidrule(lr){2-5} \cmidrule(lr){6-13}
 & \multicolumn{4}{c|}{All scenarios} & \multicolumn{4}{c}{Bike repair} & \multicolumn{4}{c}{Cooking} \\
Model & @0.1 & @0.3 & @0.5 & @0.7 & @0.1 & @0.3 & @0.5 & @0.7 & @0.1 & @0.3 & @0.5 & @0.7 \\ \hline
EgoVLPv2~\citep{pramanick2023egovlpv2egocentricvideolanguagepretraining} & 0.45 & 0.36 & 0.26 & 0.15 & 0.50 & 0.21 & \textbf{0.14} & \textbf{0.07} & 0.43 & 0.32 & 0.26 & 0.18 \\
\rowcolor{gray!20} \textbf{ViewBridge} & \textbf{0.47} & \textbf{0.38} & \textbf{0.28} & \textbf{0.17} & \textbf{0.57} & \textbf{0.29} & \textbf{0.14} & \textbf{0.07} & \textbf{0.50} & \textbf{0.39} & \textbf{0.29} & \textbf{0.19} \\
\bottomrule
\end{tabular}
\label{tab:seen_vs_unseen}
\end{table*}

\begin{table}
\centering
\caption{\textbf{Curriculum ablation on Ego-Exo4D.} We evaluate the impact of varying the length of the ego-distillation phase $l_{P}$ in our curriculum. We report $\text{Recall}@1$ with IoU at threshold $\theta$ on easy (E), moderate (M), and difficult (D) views. Exo-only distillation ($l_{P}=0\%$, row 1) yields poor performance across views. Training with a short ego-distill phase performs similarly poorly, as the model does not adapt to the new ego viewpoint. Lengthening the ego-distill phase (rows 3-4) allows sufficient adaptation to the ego viewpoint, yielding significant gains over exo-only distillation.}
\normalsize
\setlength{\tabcolsep}{3pt}
\begin{tabular}{clccc}

\multirow{5}{*}{} &  & $\theta=$0.3 & $\theta=$0.5 & $\theta=$0.7 \\ 
 & Ego-phase length ($l_{P}$) & E/M/D & E/M/D & E/M/D \\ 
\cmidrule(lr){2-5}
 & 0\% \textbf{(Exo-only)}  & .22/.22/.23 & .10/.10/.10 & .04/.03/.03\\
 & 10\% & .18/.18/.16 & .08/.08/.07 & .02/.02/.01\\
 & 25\% & .35/.36/.35 & .18/.17/.16 & .08/.08/.07\\
\rowcolor{gray!20} \cellcolor{white} &   50\% \textbf{(Ours)} & \textbf{.37}/\textbf{.36}/\textbf{.36} & \textbf{.26}/\textbf{.28}/\textbf{.28} & \textbf{.15}/\textbf{.17}/\textbf{.19}\\ 
\cmidrule(lr){2-5}
\end{tabular}
\label{tab:curr_ablations}
\end{table}

\end{document}